\documentclass[letterpaper, 10 pt, conference]{ieeeconf}  

\IEEEoverridecommandlockouts                              

\usepackage{graphics} 
\usepackage{subfigure}
\usepackage{epsfig} 
\usepackage{mathptmx} 
\usepackage{times} 
\usepackage{amsmath} 
\usepackage{amssymb}  
\usepackage{booktabs}	%
\usepackage{balance}
\usepackage{xcolor} 

\usepackage{mathtools}

\title{\LARGE \bf
PCVPC: Perception Constrained Visual Predictive Control For Agile Quadrotors
}

\author{Chao Qin$^{1}$ and Hugh H.T. Liu$^{1}$
\thanks{$^{1}$The authors are with the FSC lab, Univerisity of Toronto, Toronto, Canada
        {\tt\small chao.qin@mail.utoronto.ca, hugh.liu@utoronto.ca}}%
}

\begin{document}

\maketitle
\thispagestyle{empty}
\pagestyle{empty}

\begin{abstract}
We present a perception constrained visual predictive control (PCVPC) algorithm for quadrotors to enable aggressive flights without using any position information. Our framework leverages nonlinear model predictive control (NMPC) to formulate a constrained image-based visual servoing (IBVS) problem. The quadrotor dynamics,  image dynamics, actuation constraints, and visibility constraints are taken into account to handle quadrotor maneuvers with high agility. Two main challenges of applying IBVS to agile drones are considered: (i) high sensitivity of depths to intense orientation changes, and (ii) conflict between the visual servoing objective and action objective due to the underactuated nature. To deal with the first challenge, we parameterize a visual feature by a bearing vector and a distance, by which the depth will no longer be involved in the image dynamics. Meanwhile, we settle the conflict problem by compensating for the rotation in the future visual servoing cost using the predicted orientations of the quadrotor. Our approach in simulation shows that (i) it can work without any position information, (ii) it can achieve a maximum referebce speed of $9$ m/s in trajectory tracking without losing the target, and (iii) it can reach a landmark, e.g., a gate in drone racing, from varied initial configurations.
\end{abstract}

\section{INTRODUCTION}

The emergence of first-person-view (FPV) drones opens up a new challenge of drone racing. Due to the limited onboard sensing capability of a camera, the importance of perception awareness has been emphasized in recent unmanned aerial vehicles (UAVs) research, from planning \cite{spasojevic2020joint, murali2019perception, wang2021visibility, penin2017vision} to control \cite{falanga2018pampc, jacquet2020motor, allibert2010predictive}. For example, 
to traverse a gate safely, it is required to keep it visible at all times; and encouraging UAVs to look at texture-rich regions can effectively reduce errors in visual odometry \cite{murali2019perception}.

Most of these perception-aware methods are built upon an underlying assumption that the position of the quadrotor can be accurately measured, e.g., via GPS, motion capture systems, visual-inertial odometry (VIO), or simultaneous localization and mapping (SLAM). However, this assumption does not hold in some cases: GPS and motion capture systems cannot work in unknown indoor environments; in dynamic or feature-less scenes, VIO and SLAM will degrade seriously; in high-speed flights, the high latency of an onboard localization system is unacceptable and one needs to trade off accuracy for speed.

We seek to answer the following question. \textbf{Can we remove the dependence on the positioning system but preserve the agility of the autonomous drones?} There are two benefits from this viewpoint. First, it immunizes the controller from the deficiencies of the localization system. Second, we can free the computational resources preserved for onboard localization. 

\begin{figure}[t]
	\centering
	\framebox{\includegraphics[width=0.45\textwidth]{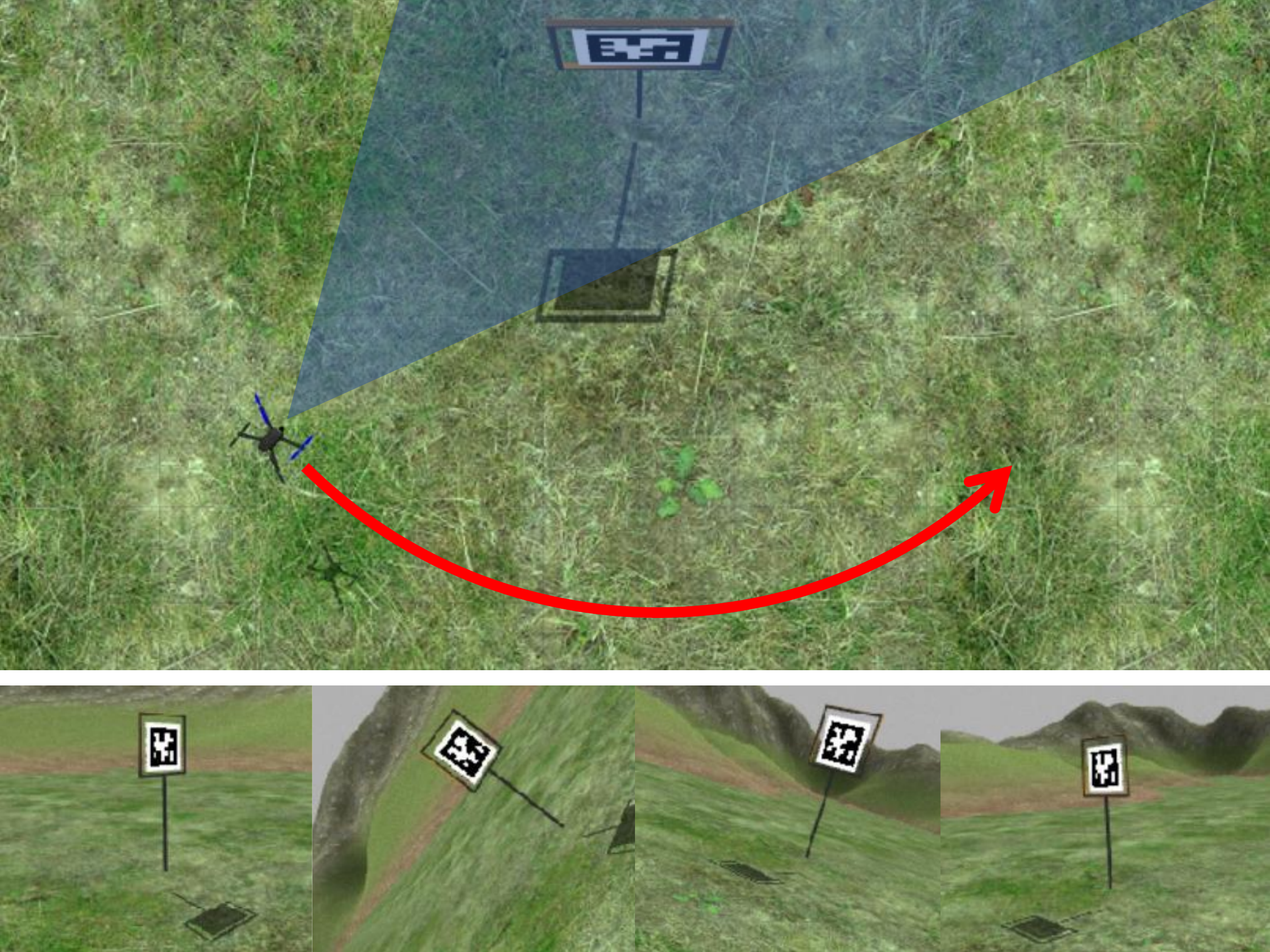}}
	\caption{An example of tracking a high-speed trajectory while keeping the visibility of the gate. The blue triangle region is the camera field of view. The lower four images were captured while the proposed method was tracking this trajectory at a maximum reference speed of $9$ m/s.}
	\label{fig_vpc}
\end{figure}

Although image-based visual servo (IBVS) does not depend on any positioning system, it fails to tackle agile underactuated robots for two reasons. First, it is difficult to handle constraints in IBVS \cite{penin2017vision}. Second, depth measurements are typically inaccurate under intense rolling and pitching, which may result in control instabilities \cite{roque2020fast}. For example, when tracking a quarter-circle trajectory with a high speed as shown in Fig. \ref{fig_vpc}, the quadrotor needs to change its orientation very frequently, making the depth hard to precisely capture. Moreover, it is infeasible for traditional IBVS methods to keep the landmark visible at such a high speed.

Our goal is to solve these two challenges of IBVS and develop an image-based high-speed drone that can work without any positioning system. The proposed method is based on two insights: (i) to fully leverage the agility of autonomous drones under an image-based control scheme, we need to consider both the quadrotor dynamics and image dynamics; and (ii) replace depth by a rotation-invariant metric, e.g., distance.

\subsection{Contributions}

In this paper, we present PCVPC, a perception constrained visual predictive control algorithm to enable agile quadrotor flights. An IBVS task is formulated as an optimal control problem to handle the quadrotor dynamics and image dynamics, as well as the actuation constraint and perception constraint, in a single framework.

To remove the depth value, we parametrize an image feature point using a bearing vector and a distance. A minimal representation of the image dynamics will be derived using 2D-manifold. Then, we model them as state variables to conduct visual predictive control.

Three objective functions are jointly optimized in our framework: (i) visual servoing objective (for minimizing the image-feature error to reach the desired configuration); (ii) perception objective (for maximizing the visibility of the landmark to improve sensing reliability); and (iii) action objective (for stabilizing quadrotors).

The main contributions of our work are as follows:
\begin{itemize}
	\item A constrained image-based visual servoing algorithm for agile quadrotors is developed;
	\item A novel visual servoing strategy using bearing vectors and distances is proposed to handle aggressive maneuvers;
	\item To our best knowledge, this is the first work that applies visual predictive control in high-speed quadrotors. The source code will be available online.
\end{itemize}

\subsection{Related Works}

With the popularity of FPV drones, there is an increasing number of works that aim to 
boost perception quality in order to improve system efficiency and robustness. In \cite{spasojevic2020joint}, Spasojevic et al. found that keeping visual features that have small area-of-search in the image can reduce the time for feature tracking and thus lower the computational burden. Furthermore, they proposed to maximize the co-visibility of features in trajectory planning to improve visual navigation quality \cite{spasojevic2020perception}. The importance of visibility in aerial tracking was also emphasized in \cite{wang2021visibility}, in which Wang et al. designed a perception cost that can be jointly optimized in trajectory optimization to achieve desired sensing reliability.

Control algorithms that consider perception quality can be categorized into position-based and image-based approaches. The key difference between these two methods is that in position-based approaches, the perception objective serves to improve navigation robustness and the task is mainly conducted by waypoint tracking, while in image-based approaches, the perception objective is used for both purposes.

In position-based approaches, PAMPC \cite{falanga2018pampc} is the first work that took perception objectives as a cost in a nonlinear model predictive control (NMPC) framework which would be jointly optimized with the action cost to meet the sensing requirements. In \cite{jacquet2020perception} and \cite{jacquet2020motor}, Jacquet et al. produced a rotor-level NMPC for both underactuated and tilted-propeller MAVs. The perception constraints were expressed as the limitations of the vertical and horizontal angles of the field of view (FoV). In addition, position-based visual servo (PBVS) also belongs to this category. In \cite{sheckells2016optimal}, Sheckells et al. introduced an optimal PBVS method that can minimize image feature reprojection error to ensure feature visibility. In \cite{potena2017effective}, Potena et al. developed a two-step PBVS method: first, an optimal global trajectory that takes into account the dynamic and perception constraints was computed; and then, an NMPC controller was applied to track the resulting trajectory.

In image-based approaches, IBVS \cite{chaumette2006visual} is the most representative one. However, 
handling constraints such as visibility limits is a tricky problem for IBVS \cite{allibert2010visual}. To deal with such problems, visual predictive control (VPC) \cite{weipeng2014predictive, durand2020visual, fusco2020integrating, zhang2021robust} was proposed which deems IBVS as a model predictive control (MPC) problem. While it is common to apply VPC in fully-actuated robots such as manipulators \cite{weipeng2014predictive} and ground vehicles \cite{durand2020visual}, it is uncommon to utilize VPC in quadrotors. One serious challenge is the coupling between the translational and rotational movements in a quadrotor. To tackle this coupling, Zhang et al. defined a virtual frame to compensate for the rolling and pitching of the quadrotor \cite{zhang2021robust}. In \cite{roque2020fast}, Roque et al. split the system into an IBVS-based velocity solver and an MPC-based velocity controller and left the underactuation  to the velocity controller. Owing to the split, the IBVS algorithm can work as a stand-alone module that does not need to consider the actual quadrotor dynamics. 

\section{Modeling}

In this section, we detail our parametrization of an image feature point, and the dynamic models of both the image feature and the quadrotor will be introduced. 

\begin{figure}[t]
	\centering
	\framebox{\includegraphics[width=0.45\textwidth]{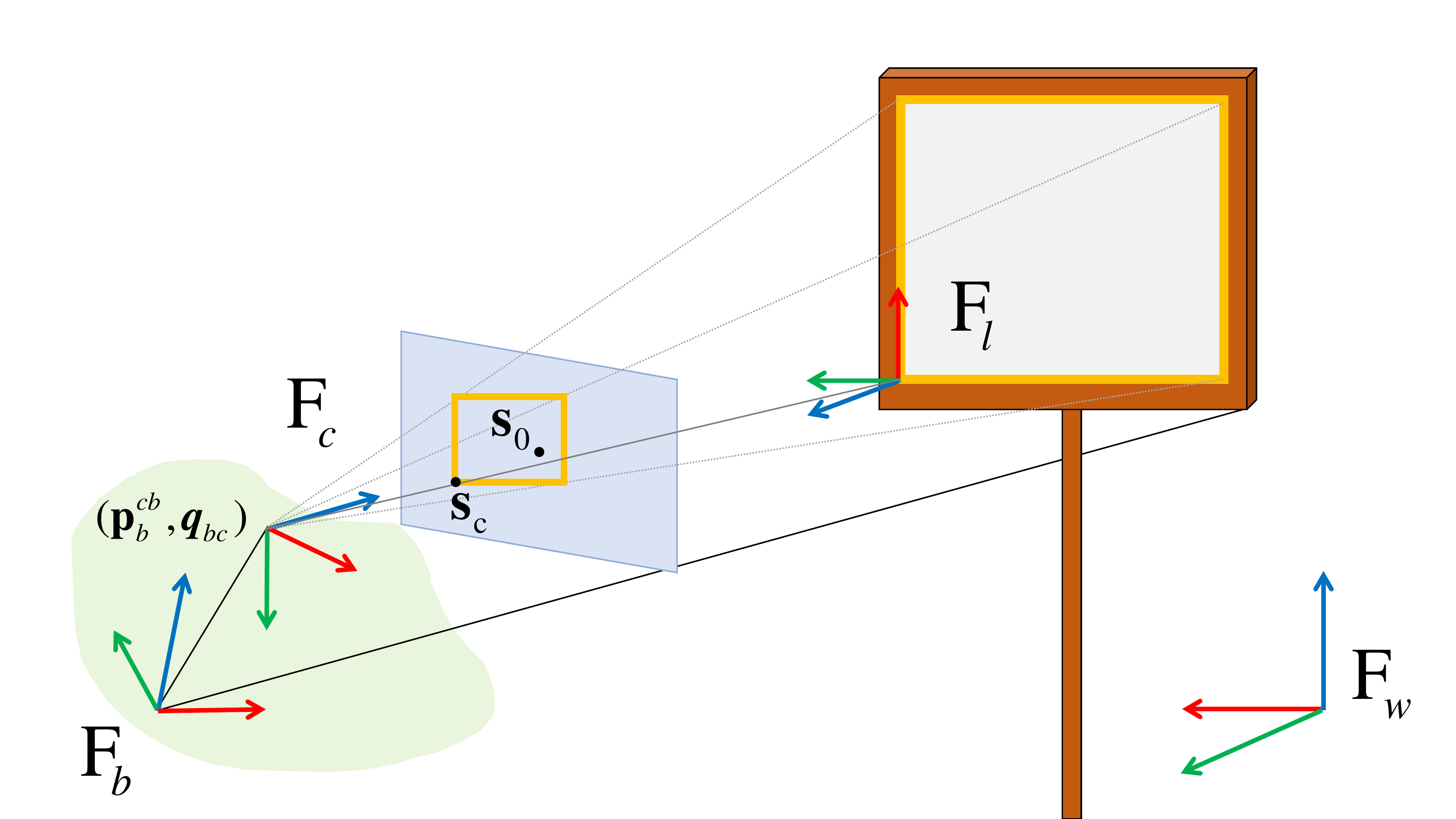}}
	\caption{A schematics representing the world frame $\mathcal{F}_{w}$, body frame $\mathcal{F}_{b}$, camera frame $\mathcal{F}_{c}$ and landmark frame $\mathcal{F}_{l}$. The standard basis is colored as \{\textcolor{red}{$\mathbf{x}$}, \textcolor{green}{$\mathbf{y}$}, \textcolor{blue}{$\mathbf{z}$}\}. The relative pose from the camera frame to the body frame is represented by $(\mathbf{p}_{b}^{cb},\mathbf{q}_{bc})$. A landmark centerd at the origin of $\mathcal{F}_{l}$ is projected onto image coordinate $\mathbf{s}_c$. The image center is denoted by $\mathbf{s}_0$.}
	\label{fig_frames}
\end{figure}

\subsection{Notation Definition}

We define the world frame as $\mathcal{F}_{w}$, the body frame as $\mathcal{F}_{b}$, the camera frame as $\mathcal{F}_{c}$, and the landmark frame as $\mathcal{F}_{l}$. A clear overview of the reference frames is depicted in Figure \ref{fig_frames}. The position of the landmark frame with respect to the camera frame, expressed in the camera frame is denoted by $\mathbf{p}_{c}^{lc}=(x_{c},y_{c},z_{c})^{T}$. We use quaternions to represent the orientation between frames. For example, $\mathbf{q}_{cl}$ represents the rotation from the landmark frame to the camera frame. An operator $\otimes$ is used to denote the Hamilton product of two quaternions, and an operator $\odot$ is used to express the multiplication between a quaternion and a vector. For example, rotating the landmark position from the camera frame to the world frame can be written as $\mathbf{p}_{w}^{lc}=(\mathbf{q}_{wb}\otimes\mathbf{q}_{bc})\odot\mathbf{p}_{c}^{lc}$.

\subsection{Bearing Vector}

A bearing vector is a 3D unit vector on 2-sphere manifold. To yield simple analytical derivatives without having the "hairy ball theorem" issue, we employ a quaternion, $\mathbf{q}_{cl}$, as the underlying representation for the bearing vector of a landmark as in \cite{bloesch2017iterated}. Two functions involved in the bearing vector transformation are defined below:
\begin{align}
	n(\mathbf{q}_{cl})&=\mathbf{q}_{cl}\odot \mathbf{e}_{z}\subset \mathbb{R}^{3} \label{equ_bv_1}, \\
	N(\mathbf{q}_{cl})&=[\mathbf{q}_{cl}\odot \mathbf{e}_{x},\mathbf{q}_{cl} \odot \mathbf{e}_{y}]\in\mathbb{R}^{3\times2}, \label{equ_bv_2}
\end{align}
where $n(\mathbf{q}_{cl})$ returns the original 3D unit vector and $N(\mathbf{q}_{cl})$ linearly projects a 3D vector onto the 2D tangent space around the bearing vector; $(\mathbf{e}_x, \mathbf{e}_y, \mathbf{e}_z)$ is the standard basis for the 3D space. For better understandings, we can think of Equations  (\ref{equ_bv_1}) and (\ref{equ_bv_2}) as describing the standard basis of the landmark frame with respect to the camera frame. Detailed definiton of bearing vectors can be found in \cite{bloesch2017iterated}.

We use an operator $\left[\cdot\right]_{z}$ to transform a 3D vector to homogeneous coordinates. For example, the image coordinate of a feature can be represented by $\mathbf{s}_{c}=\left[\mathbf{p}_{c}^{lc}\right]_{z}=(x_{c}/z_{c},y_{c}/z_{c},1)^{T}$. Note that to simplify notations, in this paper, the image coordinate refers to homogeneous coordinates instead of pixel coordinates. The following relasionship exists for the a feature point in the image:
\begin{equation}
\mathbf{p}_{c}^{lc}=\mathbf{s}_{c} \cdot z_{c} = n(\mathbf{q}_{cl}) \cdot d, \label{equ_bv_hc}
\end{equation}
where $z_{c}$ denotes the depth and $d$ the distance to the feature.
 
We can transform a bearing vector to the corresponding image coordinate, and vice versa. To calculate $\mathbf{q}_{cl}$ from $\mathbf{s}_{c}$, three steps are needed. First, we normalize $\mathbf{s}_{c}$. Second, the angle-axis vector between the normalized $\mathbf{s}_{c}$ and $\mathbf{e}_z$ is calculated using function $\theta(\mathbf{a},\mathbf{b})=\frac{\arccos(\mathbf{b}^{T}\mathbf{a})}{\mathbf{b}\times\mathbf{a}}\mathbf{b}\times\mathbf{a}$, where $\mathbf{a}$, $\mathbf{b}$ are two unit vectors. Finally, we represent the angle-axis vector as a rotation to yield $\mathbf{q}_{cl}$. By comparison, it is much easier to obtain $\mathbf{s}_{c}$ from $\mathbf{q}_{cl}$, which is via $\mathbf{s}_{c}=\left[n(\mathbf{q}_{cl})\right]_{z}$. 

\subsection{Quadrotor Dynamics}

Let $\mathbf{v}_{w}$ and $\mathbf{q}_{wb}$ be the velocity and the orientation of the body frame with respect to the world frame. Additionally, let $\boldsymbol{\omega}_{b}$ be the angular velocity expressed in the body frame, and $\mathbf{c}=(0,0,c)^{T}$ the mass-normalized thrust vector. The quadrotor dynamics without position can be modeled as:
\begin{align}
	\begin{split}
		\mathbf{\dot{v}}_{w}&=\mathbf{q}_{wb}\odot\mathbf{c}+\mathbf{g}_{w},\\
		\dot{\mathbf{q}}_{wb}&=\frac{1}{2}\left[\begin{array}{c}
			0\\
			\boldsymbol{\omega}_{b}
		\end{array}\right]\otimes\mathbf{q}_{wb},
	\end{split}\label{equ_quad_dynamics}
\end{align}
where $\mathbf{g}_{w}=(0,0,-9.81)^{T}$ is the gravity vector.

\subsection{Image Feature Parametrization and Dynamics}

Recently researches discover that human drone racing pilots focus their eye gaze onto one specific point on the gates around $1.5$ sec and $16$ meters before the drone traverses the gates \cite{pfeiffer2021human}. Inspired by this finding, our work uses only one fixed feature to mimic the gaze fixation.

Traditional visual servoing approaches express an image feature using homogeneous coordinates. The advantage of this parametrization is that only a minimum number of states are required for visual prediction as in \cite{fusco2020integrating}. However, it brings a disadvantage that the image dynamics and the system state will contain depths, which, if inaccurate, may jeopardize visual servoing stability \cite{roque2020fast}. Furthermore, depths are subject to both relative positions and orientations between the quadrotor and the landmark, and a minor change of orientation may result in a large depth change. As a consequence, it is hard to measure the true depth when the quadrotor performs aggressive flights.

By comparison, distance is a much stable metric since it is invariant to orientations. However, it makes the image dynamics much complicated if we stick to homogeneous coordinats. Moreover, it is inefficient to recover $\mathbf{p}_{c}^{lc}$ using a $d$ and $\mathbf{s}_{c}$ since we need to normalize $\mathbf{s}_{c}$ first. To solve these problems, it is necessary to consider a new parametrization and new image dynamics that are compatible with the distance metric.

In this paper, a 3D feature is parametrized by a distance $d$ and a bearing vector represented by a quaternion $\mathbf{q}_{cl}$. Based on this formulation, we are able to compute $\mathbf{p}_{c}^{lc}$ using pure multiplications as in Equation (\ref{equ_bv_hc}), and obtain a much simpler image dynamics:
\begin{align}
\begin{split}
\dot{\mathbf{q}}_{cl}&=\frac{1}{2}N(\mathbf{q}_{cl})^{T}\left(-\boldsymbol{\omega}_{c}-n(\mathbf{q}_{cl})^{\times}\frac{\mathbf{v}_{c}}{d}\right),\\
\dot{d}&=-n(\mathbf{q}_{cl})^{T}\mathbf{v}_{c},
\end{split} \label{equ_img_dynamics}
\end{align}
where the skew operator $(\cdot)^{\times}$ produces the cross-product matrix of a 3D vector:
\begin{equation}
\mathbf{a}^{\times}=\left[\begin{array}{ccc}
	0 & -a_{z} & a_{y}\\
	a_{z} & 0 & -a_{x}\\
	-a_{y} & a_{x} & 0
\end{array}\right].
\end{equation}
$\mathbf{v}_{c}$ and $\boldsymbol{\omega}_c$ denote the linear velocity and angular velocity represented in the camera frame, respectively. The transformations from $\mathbf{v}_{w}$ to  $\mathbf{v}_{c}$ and $\boldsymbol{\omega}_{b}$ to $\boldsymbol{\omega}_{c}$ are formulated as:
\begin{align}
	\begin{split}
\mathbf{v}_{c}&=\mathbf{q}_{bc}^{-1}\odot(\mathbf{q}_{wb}^{-1}\odot\mathbf{v}_{w}+\boldsymbol{\omega}_{b}^{\times}\mathbf{p}_{b}^{cb}),\\
\boldsymbol{\omega}_{c}&=\mathbf{q}_{bc}^{-1}\odot\boldsymbol{\omega}_{b}.
	\end{split}.
\end{align}
Note that although there is a redundancy in the representation of a 3D feature point, no extra constraints will be introduced into optimization.

We provide a detailed derivation of Equation (\ref{equ_img_dynamics}) in the Appendix. Our contribution is a world-centric version of the bearing vector dynamics, while \cite{bloesch2015robust} only provides a robot-centric version which is not suitable for our formulation.

Finally, the state and the input vectors of the system are defined as:
\begin{align}
\mathbf{x}&=[\mathbf{v}_{w}^{T},\mathbf{q}_{wb}^{T},\mathbf{q}_{cl}^{T},d]^{T} \in\mathbb{R}^{12}, \\
\mathbf{u}&=[c,\boldsymbol{\omega}_{b}^{T}]^{T} \in\mathbb{R}^{4}. \label{equ_states} 
\end{align}

\section{Optimal Control Problem Formulation}

The typical problem tackled by PCVPC is to reach a landmark as fast as possible while maintaining visibility on the landmark. One can generalize PCVPC to complex scenarios by choosing a sequence of landmarks, e.g., racing gates, and reaching them one by one.

In this section, we introduce two constraints and three objective functions applied to PCVPC. An optimal control problem will be formulated to compute the desired control inputs.

\subsection{System Constraints}

\subsubsection{Perception Constraint}
Preventing features from leaving the camera FoV is of paramount importance. We design a hard visibility constraint in the system to satisfy this requirement. Thanks to the bearing vector, we are able to express the visibility constraint without using external parameters:
\begin{equation}
	\mathbf{s}_{min}\leq\mathbf{s}_{c}\leq\mathbf{s}_{max}, \label{equ_px_constraints}
\end{equation}
where $\mathbf{s}_{c}=\left[n(\mathbf{q}_{cl})\right]_{z}$.

\subsubsection{Actuation Constraint}
The control inputs $\mathbf{u}$ in Equation (\ref{equ_states}) are limited by the following constraints:
\begin{align}
\begin{split}
c_{min}\leq &c\leq c_{max}, \\
\boldsymbol{\omega}_{min}\leq&\boldsymbol{\omega}_{b}\leq\boldsymbol{\omega}_{max}. 
\end{split} \label{equ_input_constraints}
\end{align}

\subsection{Objective Functions}

\subsubsection{Visual Servoing Objective}
The visual servoing objective aims at minimizing the errors in terms of image coordinates and distances, between reference states and current states. We define the visual servoing objective function at each time step as
\begin{equation}
\mathcal{L}_{vs}(\mathbf{x},\mathbf{u})=\Vert\mathbf{s}_{c^{*}}-\mathbf{s}^{*}\Vert_{\mathbf{Q}_{s}}^{2}+\Vert d-d^{*}\Vert_{Q_{d}}^{2},\label{equ_vs_cost}
\end{equation}
where $\Vert \cdot \Vert^2_{\mathbf{Q}}$ denotes the weighted squared Euclidean norm; $\mathbf{Q}_{s}$ is the diagonal weight matrix for image errors; $Q_{d}$ is the scalar weight for distance errors; $d^{*}$ and $d$ are reference distance and current distance, respectively; $\mathbf{s}^{*}$ denotes the reference image coordinate while $\mathbf{s}_{c^{*}}$ denotes the current image coordinate after rotational compensation:
\begin{equation}
\mathbf{s}_{c^{*}}=\left[n((\mathbf{q}_{bc}^{-1}\otimes\mathbf{q}_{wb}\otimes\mathbf{q}_{bc}\otimes\mathbf{q}_{cl})\right]_{z}. \label{equ_rot_compensation}
\end{equation}

We should notice that by tracking an image coordinate and a distance together, the controller is indeed doing local position tracking. Therefore, we can transform relative positions into reference image coordinates and distances to enable trajectory tracking.

\subsubsection{Perception Objective}
The perception objective aims to maximize the visibility of the landmark by driving the feature to the image center. The objective function can be described as:
\begin{equation}
\mathcal{L}_{p}(\mathbf{x},\mathbf{u})=\Vert\mathbf{s}_{c}\Vert_{\mathbf{Q}_{p}}^{2}, \label{equ_perception_cost}
\end{equation}
where $\mathbf{s}_{c}=\left[n(\mathbf{q}_{cl})\right]_{z}$ is the image coordiante of the perceived feature and $\mathbf{Q}_{p}$ is the corresponding diagonal weight matrix.

\subsubsection{Action Objective}

The action objective is essential to stabilize a quadrotor, which consists of velocity tracking and orientation (i.e., roll, pitch and heading) tracking:
\begin{equation}
\mathcal{L}_{a}(\mathbf{x},\mathbf{u})=\Vert\mathbf{v}_{w}-\mathbf{v}_{w}^{*}\Vert_{\mathbf{Q}_{v}}^{2}+\Vert\mathbf{q}_{wb}-\mathbf{q}_{wb}^{*}\Vert_{\mathbf{Q}_{q}}^{2}, \label{equ_action_cost}
\end{equation}
where $\mathbf{Q}_{v}$ and $\mathbf{Q}_{q}$ are diagonal weight matrices for velocity and orientation errors, respectively.

\subsection{Optimal Control Problem}

The non-linear optimization problem can be formulated as:
\begin{align}
	\begin{split}
		\min_{\mathbf{u}} \quad & \int_{t_{0}}^{t_{f}}\mathcal{L}_{vs}(\mathbf{x},\mathbf{u})+\mathcal{L}_{p}(\mathbf{x},\mathbf{u})+\mathcal{L}_{a}(\mathbf{x},\mathbf{u})dt\\
		\textrm{s.t.} \quad & \dot{\mathbf{x}}=f(\mathbf{x},\mathbf{u}),\\
		&\mathbf{s}_{min}\leq\mathbf{s}_{c}\leq\mathbf{s}_{max},\\
		&c_{min}\leq c\leq c_{max},\\
		&\boldsymbol{\omega}_{min}\leq\boldsymbol{\omega}_{b}\leq\boldsymbol{\omega}_{max},
	\end{split}\label{equ_mpc}
\end{align}
where $t_{0}$ is the start time and $t_{f}$ is the end time. The system dynamics $\dot{\mathbf{x}} = f(\mathbf{x}, \mathbf{u})$ can obtained via the concatenation of Equations (\ref{equ_quad_dynamics}) and (\ref{equ_img_dynamics}); the constraints are defined in Equations (\ref{equ_px_constraints}) and (\ref{equ_input_constraints}); and the objective functions can be found in Equations (\ref{equ_vs_cost}), (\ref{equ_perception_cost}), and (\ref{equ_action_cost}).

We apply multiple shooting with $N$ samples as a transcription method and the Runge-Kutta method for integration. To account for the drifts in visual prediction, our framework works in a receding-horizon fashion by iteratively solving the above optimization problem, with only the first control input being executed.

\section{Simulation Results}

We present realistic simulation results using Gazebo and the Pixhawk4 controller \cite{pixhawk42021controller} with software-in-the-loop (SITL) simulation. Our goal is to validate the proposed method in varied scenarios and show its potentials in high-speed applications with agile maneuvers.

\subsection{Simulation Setup}

\begin{figure}[b]
	\centering\
	\includegraphics[width=0.45\textwidth]{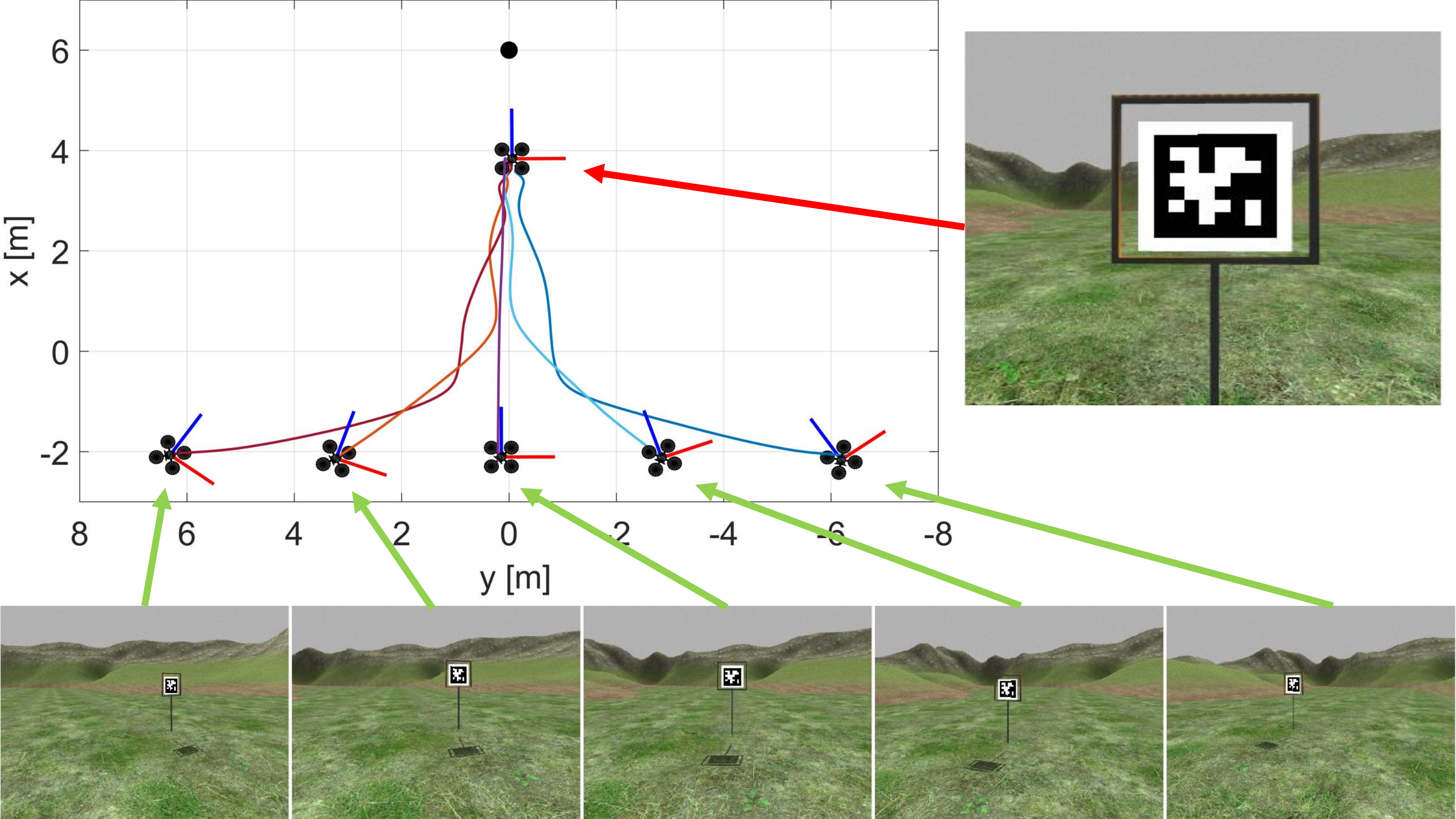}
	\caption{Five quadrotor paths from five different initial poses to the same goal. Five initial images and the reference image are captured.}
	\label{fig_racing}	
\end{figure}

The system is implemented in C++ and runs in ROS environment. Our NMPC framework is based on ACADO \cite{houska2011acado} with qpOASES as the solver. We set the time step as $dt = 0.5$ sec with a horizon of $N = 20$. After each iteration, we put all the computed control inputs in a queue and publish them at $20$ Hz in another loop. A dynamic $\mathbf{Q}_{vs}$ is applied for the visual servoing objective function to compensate the magnitude inconsistency between distances $d$ and image coordinates $\mathbf{s}_c$. The weight for each state can be found in our open-sourced code.

We simulated a 3DR Iris quadrotor with a forward-looking depth camera. The Pixhawk4 controller was configured into the offboard mode to enable thrust and body rate control. As shown in Fig. \ref{fig_racing}, the landmark is a rectangular gate with a fiducial marker \cite{wang2016apriltag} inside. The target feature is selected as the center of the fiducial marker, whose position is $\mathbf{p}_{w}^{lw}=(6,0,3)^T$ for all simulations. Similar to \cite{jacquet2020motor}, the decision of using fiducial markers comes from the emphasis of this work on designing a generic perception-aware control framework instead of image processing, and thus the feature extraction part can be suitably abstracted. Users can replace the marker by their own landmarks to accomplish different tasks.

In terms of state measurements, we assume that the velocity and orientation (i.e., roll, pitch, and heading) of the quadrotor can be accurately measured. For a real quadrotor that equipped with a PX4FLOW \cite{honegger2013open} and IMUs, this is a reasonable assumption. Furthremore, we assume that the size of the fiducial marker is given as a prior and we can estimate the distance from the robot to the marker center using the perspective-n-point (PnP) method \cite{lepetit2009epnp}. Additionally, the camera intrinsic/extrinsic parameters are assumed to be well-calibrated.

Regarding how to obtain the reference image coordinate and distance, if it is a target reaching/tracking task, the image center and a small distance can be considered as the reference; if it is a trajectory tracking task, we need to transform waypoints into reference states using a camera model; and if a reference image is given as in IBVS, we can directly obtain the reference via feature extraction.

\subsection{Scenario Description and Results}

To demonstrate the effectiveness of our PCVPC, we ran three different scenarios. In the first scenario, the controller reached the gate from varied positions. In the second scenario, the controller tracked a quarter-circle trajectory with different speeds. In the third scenario, the controller is evaluated with and without perception objectives under extreme conditions.

\subsubsection{Gate-reaching Flights}

\begin{table}[htbp]
	\caption{Initial conditions of gate-reaching flights}
	\label{tab_gate_reaching}
	\centering
	\begin{tabular}{llllll}
		\hline
		No.     & 1       & 2       & 3       & 4       & 5       \\ \hline
		Position (m)   & (-2,6,3) & (-2,3,3) & (-2,0,3) & (-2,-3,3) & (-2,-6,3) \\
		Heading ($^{\circ}$) & -30      & -15      & 0       & 15     & 30     \\ \hline
	\end{tabular}
\end{table}

\begin{figure}[t]
	\centering
	\subfigure[]{\includegraphics[width=0.225\textwidth]{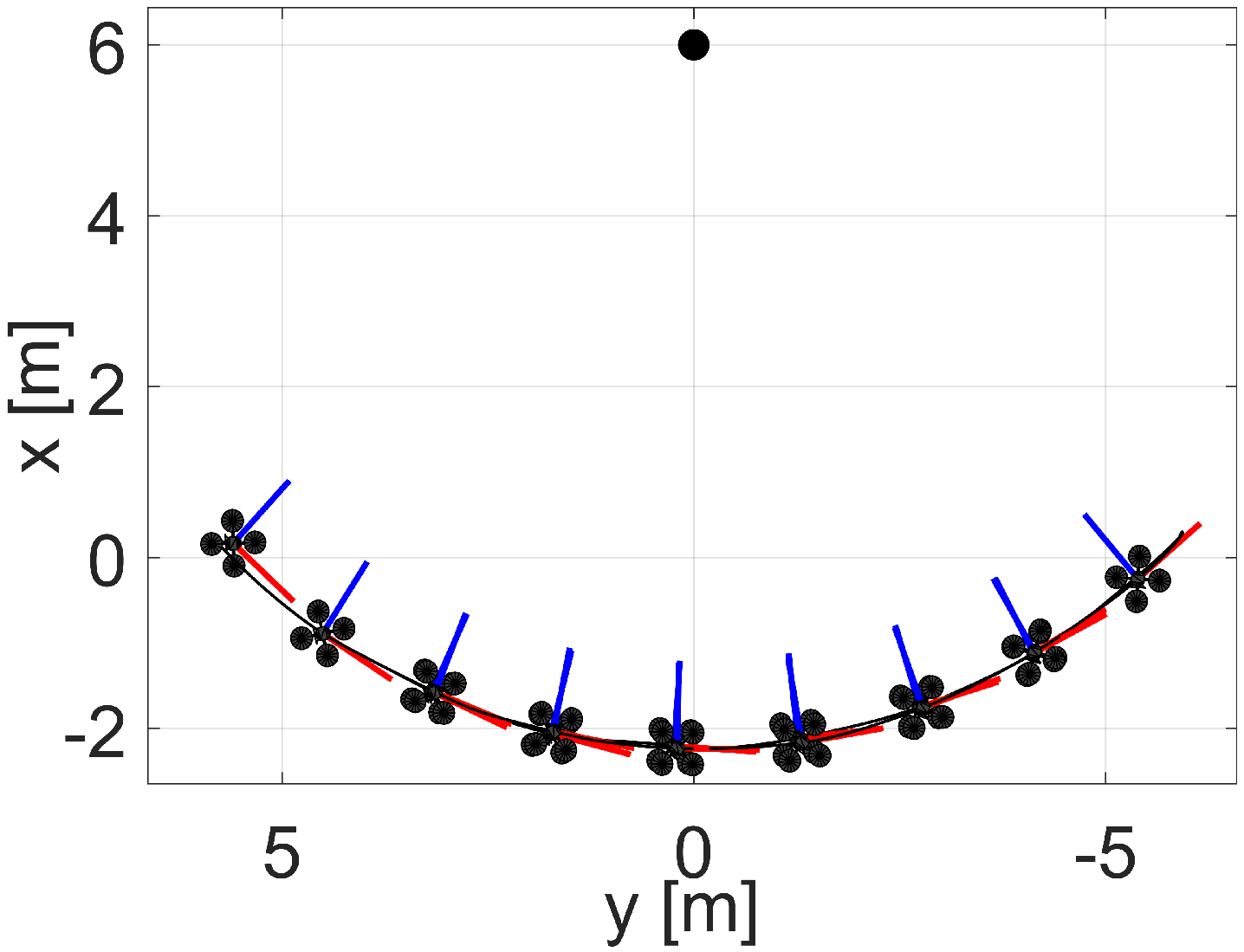}\label{fig_circle_c8_v1_xy}}
	\subfigure[]{\includegraphics[width=0.225\textwidth]{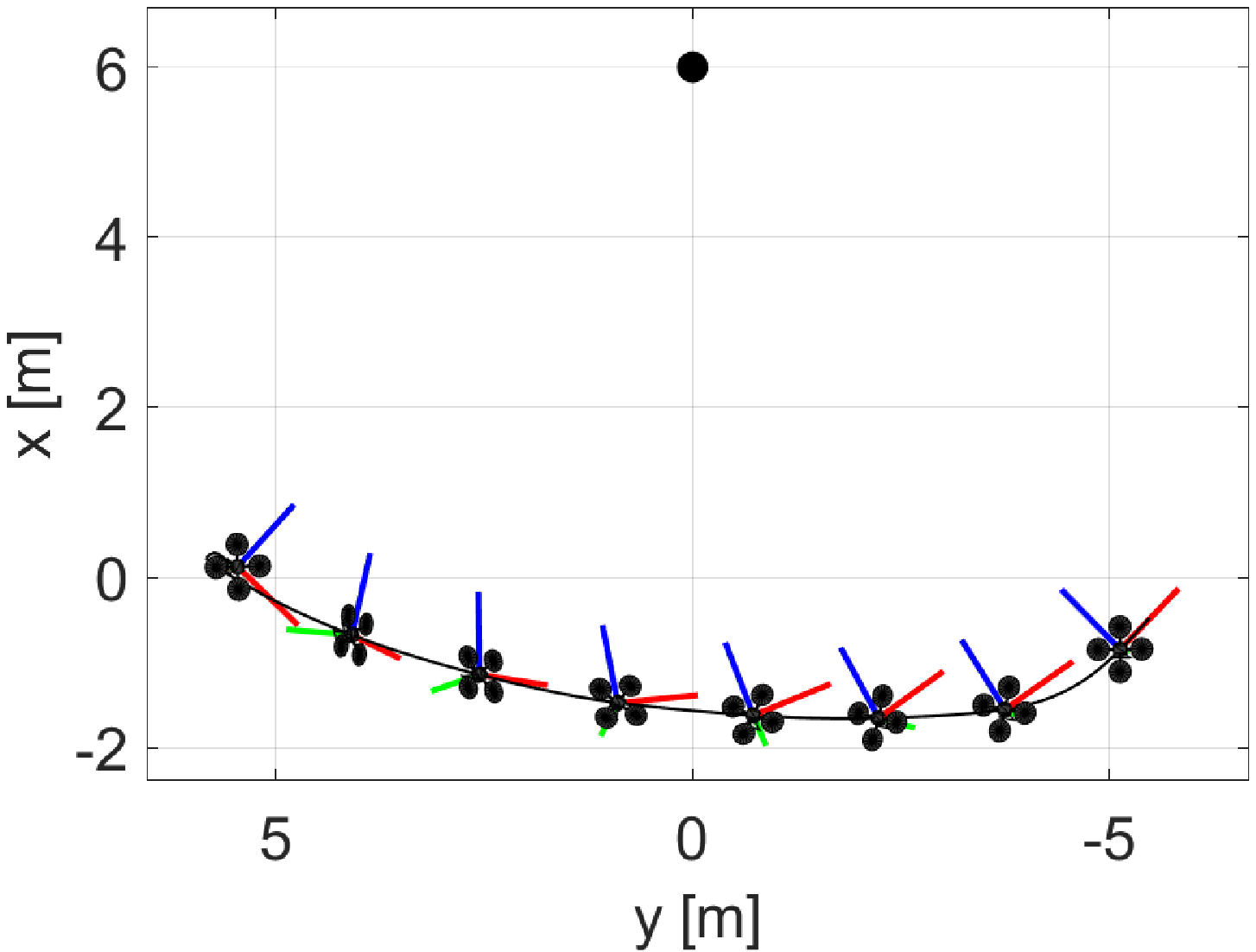}\label{fig_circle_c8_v9_xy}}
	\subfigure[]{\includegraphics[width=0.45\textwidth]{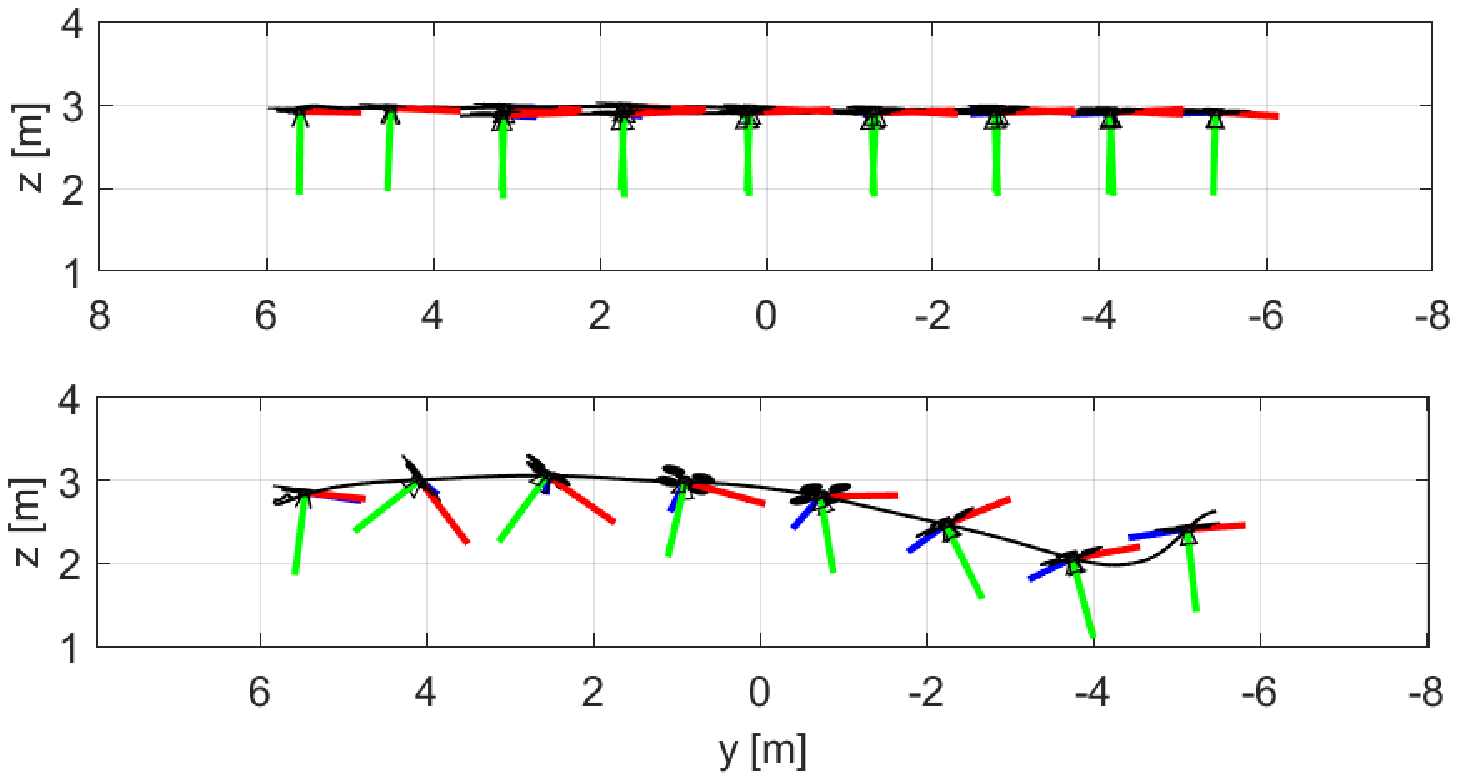}\label{fig_circle_c8_v1v9_yz}}	
	\caption{Quadrotor path in quarter-circle tracking, with the camera frame indicated by \{\textcolor{red}{$\mathbf{x}_c$}, \textcolor{green}{$\mathbf{y}_c$}, \textcolor{blue}{$\mathbf{z}_c$}\}. (a) $1$ m/s. (b) $9$ m/s. (c) comparison of altitude changes of $1$ m/s (upper) and $9$ m/s (lower).}	
	\label{fig_circle_c8_v1v9}
\end{figure}

\begin{figure}[t]
	\centering\
	\includegraphics[width=0.3\textwidth]{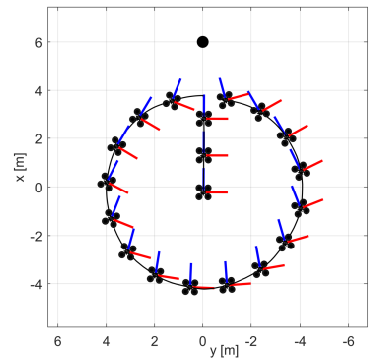}
	\caption{Quadrotor adjusted its heading to improve target visibility while executing the trajectory.}
	\label{fig_visibility}
\end{figure}

This simulation aims to show that PCVPC can converge to the reference image from different initial conditions. Table \ref{tab_gate_reaching} lists the initial poses and Fig. \ref{fig_racing} showcases the initial images and the goal image, as well as the tracking trajectories. The results show that the controller can track the reference image from all initial configurations. The average velocity in these five tasks is $2.4$ m/s. However, the velocity along each axis is quite different. From Fig. \ref{fig_racing}, we observe that the position errors in the y-axis converge much faster than those in the x-axis. That means the quadrotor flew much aggressively in rolling than in pitching. One reason is that when the gate is around the image center, conducting large rolling will not lose the gate, whereas conducting pitching to accelerate the drone forward can easily move the gate to the image upper border and then lose it. As a result, the controller will prevent large forward acceleration to maintain visibility, which also limits the forward speed.

\subsubsection{Quarter-circle Flights} \label{section_qc_flights}

This simulation aims to prove that the proposed method can enable high-speed flights and aggressive maneuvers. A horizontal quarter-circle centered at $(6,0,3)$ with a radius of $8$ m was planned as the reference trajectory. We transformed the trajectory points into reference features using a camera model and sent them to the robot along with reference velocities and headings. The performance of our framework under varied speeds is evaluated.

The results of the one running at the minimum reference speed $1$ m/s, and the one running at the maximum reference speed $9$ m/s are plot in Fig. \ref{fig_circle_c8_v1v9}. We see that agile maneuvers that require large deviations from the hover conditions are conducted to accomplish the task. From Fig. \ref{fig_circle_c8_v1v9_yz}, we also notice that there is a salient altitude change up to $1.072$ m when the robot was performing high-speed flights. The results reveal that PCVPC guarantees target visibility via adjusting its altitude during acceleration and deceleration stages.

\begin{figure}[htbp]
	\centering
	\subfigure[]{\includegraphics[width=0.225\textwidth]{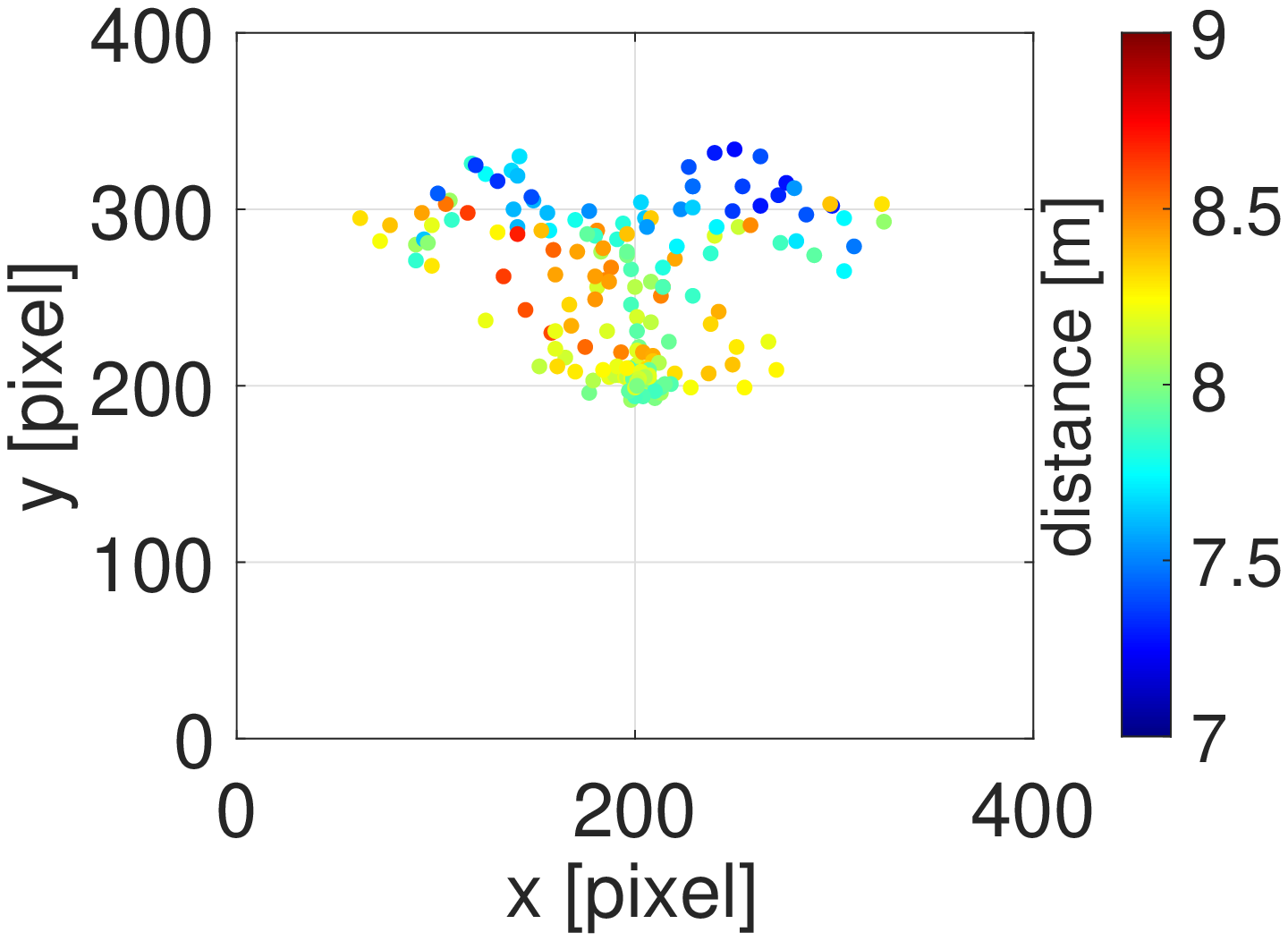}\label{fig_image_c8_v9_with}}
	\subfigure[]{\includegraphics[width=0.225\textwidth]{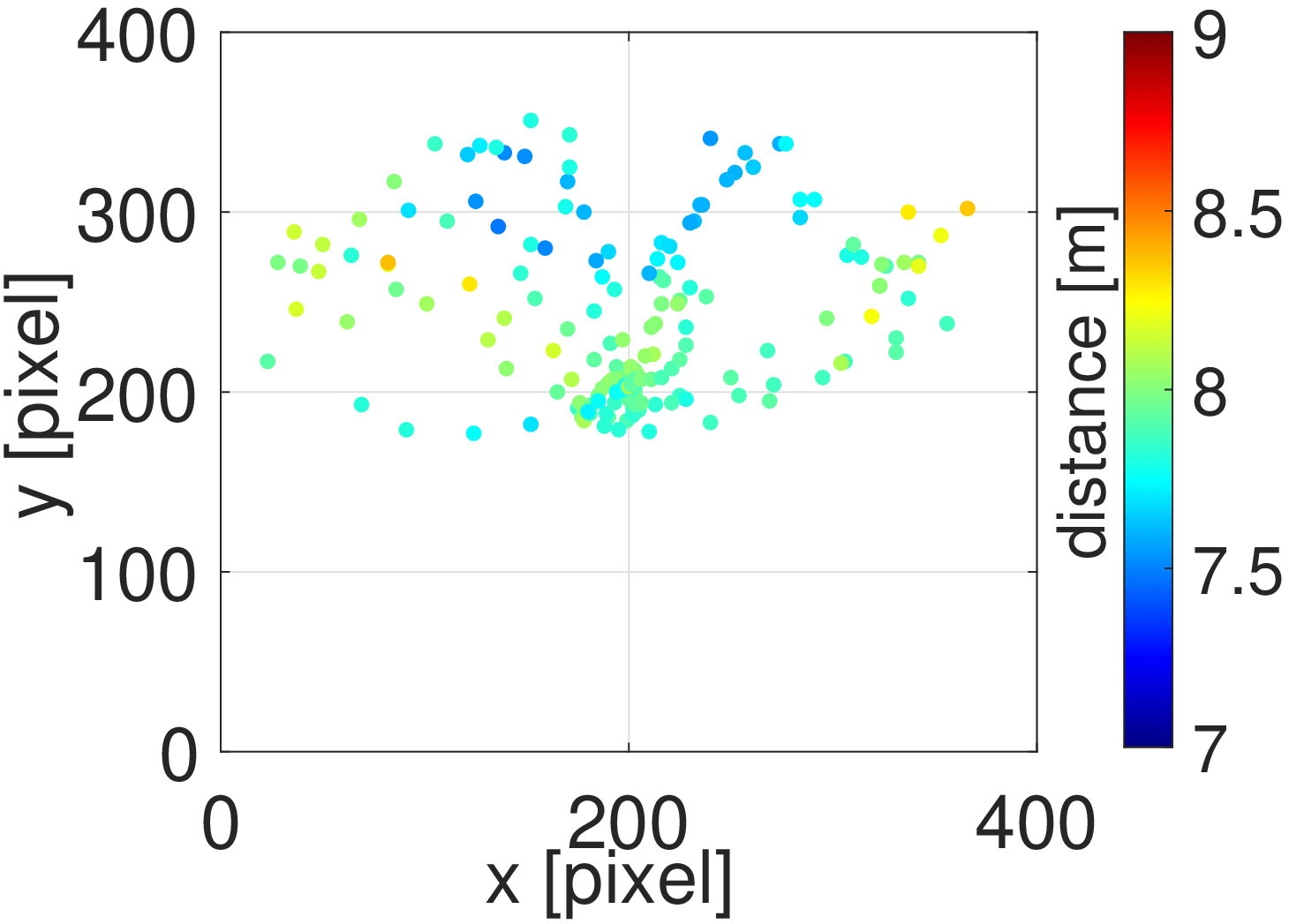}\label{fig_image_c8_v8_without}}
	\caption{Feature distribution in the image space during the tracking task. Points are colored according to the distance. Note that without perception objectives, the algorithm cannot work at $9$ m/s. (a) $9$ m/s with perception objectives. (b) $8$ m/s without perception objectives.}
	\label{fig_image_cmp}
\end{figure}

\subsubsection{Flights With and Without Perception Awareness}

\begin{table}[htbp]
	\caption{Success rates with and without perception objectives}
	\label{tab_success_rate}
	\centering
	\begin{tabular}{@{}llllll@{}}
		\toprule
		Maximum Ref. Speed (m/s)  & 6.0     & 7.0     & 8.0     & 9.0    & 10.0  \\ \midrule
		Without Perception Obj. & 100\% & 100\% & 80\%  & 0\%  & 0\% \\
		With Perception Obj.   & 100\% & 100\% & 100\% & 85\% & 0\% \\ \bottomrule
	\end{tabular}
\end{table}

We demonstrate the ability and importance of the perception objective to modulate the task in order to ensure stability.

A horizontal circle trajectory centered at $(0,0,3)$ with a radius of $4$ m was generated, and the robot was commanded to track the circle with a constant zero heading. From Fig. \ref{fig_visibility}, we see that the quadrotor was struggling to look toward to gate, keeping a non-zero heading.

To prove the importance of the perception objective, we compared the success rate of PCVPC with and without perception objectives under different speeds. The simulation setup was the same as Section \ref{section_qc_flights}. Tracking the trajectory at the given speed three times without failure was deemed as a success. For each speed, we performed $20$ simulations and recorded the success rate. The results are presented in Table \ref{tab_success_rate}. We see that adding the perception objective can effectively boost the success rate in high-speed scenarios:  without perception objectives, the success rate of PCVPC decreased dramatically from $80$ \% to $0$ \% in the $9$ m/s case; but after adding the perception objective, we can still maintain a success rate of $85$ \%.

Fig. \ref{fig_image_cmp} displays the feature distribution in the image space at their respective maximum speeds. It is clear to see that the features in Fig. \ref{fig_image_c8_v9_with} are more concentrated to the image center due to the perception objective, even under a faster speed. As shown in Fig. \ref{fig_image_c8_v8_without}, without perception objectives, it is more likely for features to approach the image border due to large rolling and pitching, adding the risks of feature loss. Therefore, it is necessary to add the perception objective to handle aggressive movements.



\subsection{Comparison of Two Visual Prediction Strategies}

\begin{figure}[htbp]
	\centering
	\subfigure[]{\includegraphics[width=0.225\textwidth]{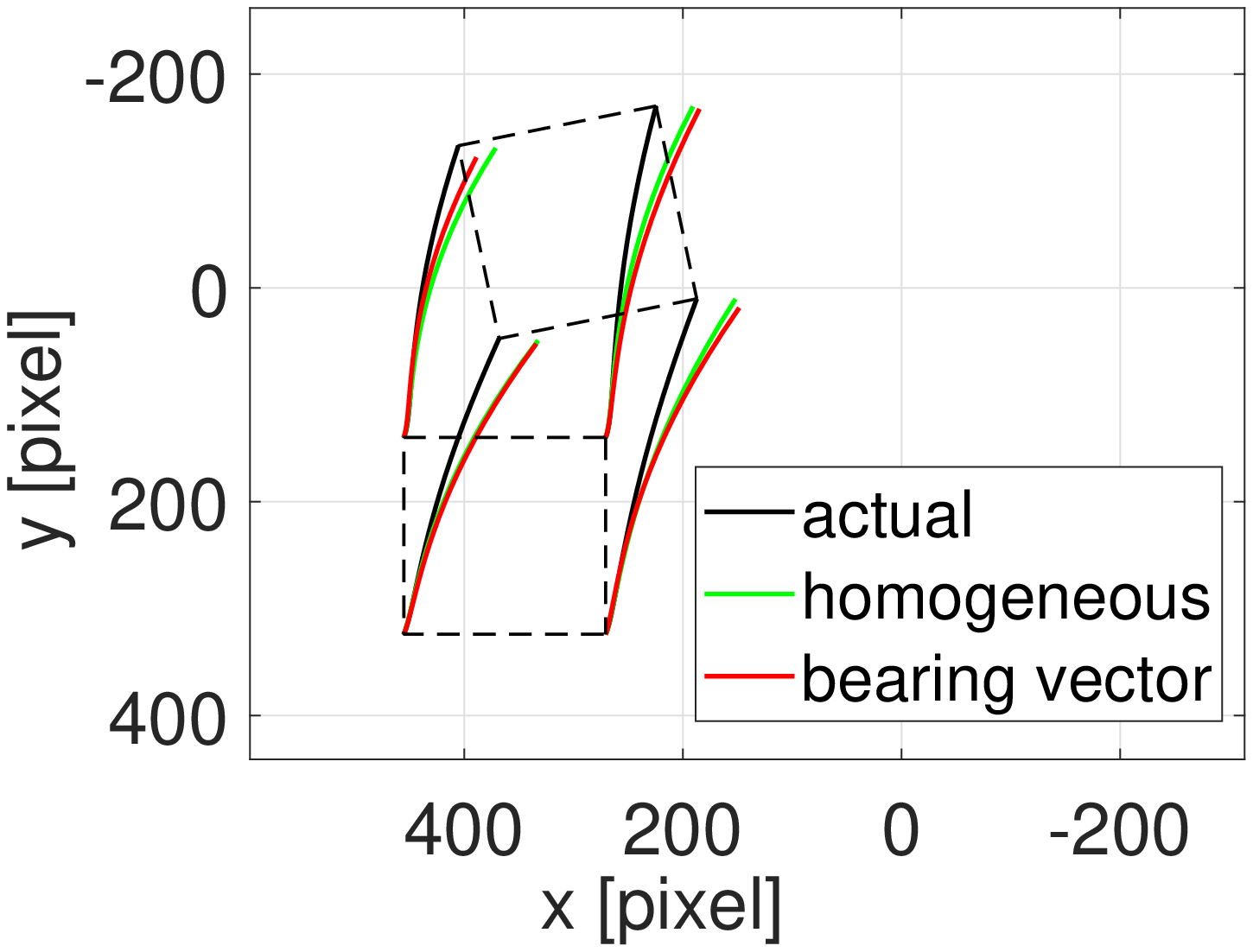}\label{fig_visual_prediction_02}}
	\subfigure[]{\includegraphics[width=0.225\textwidth]{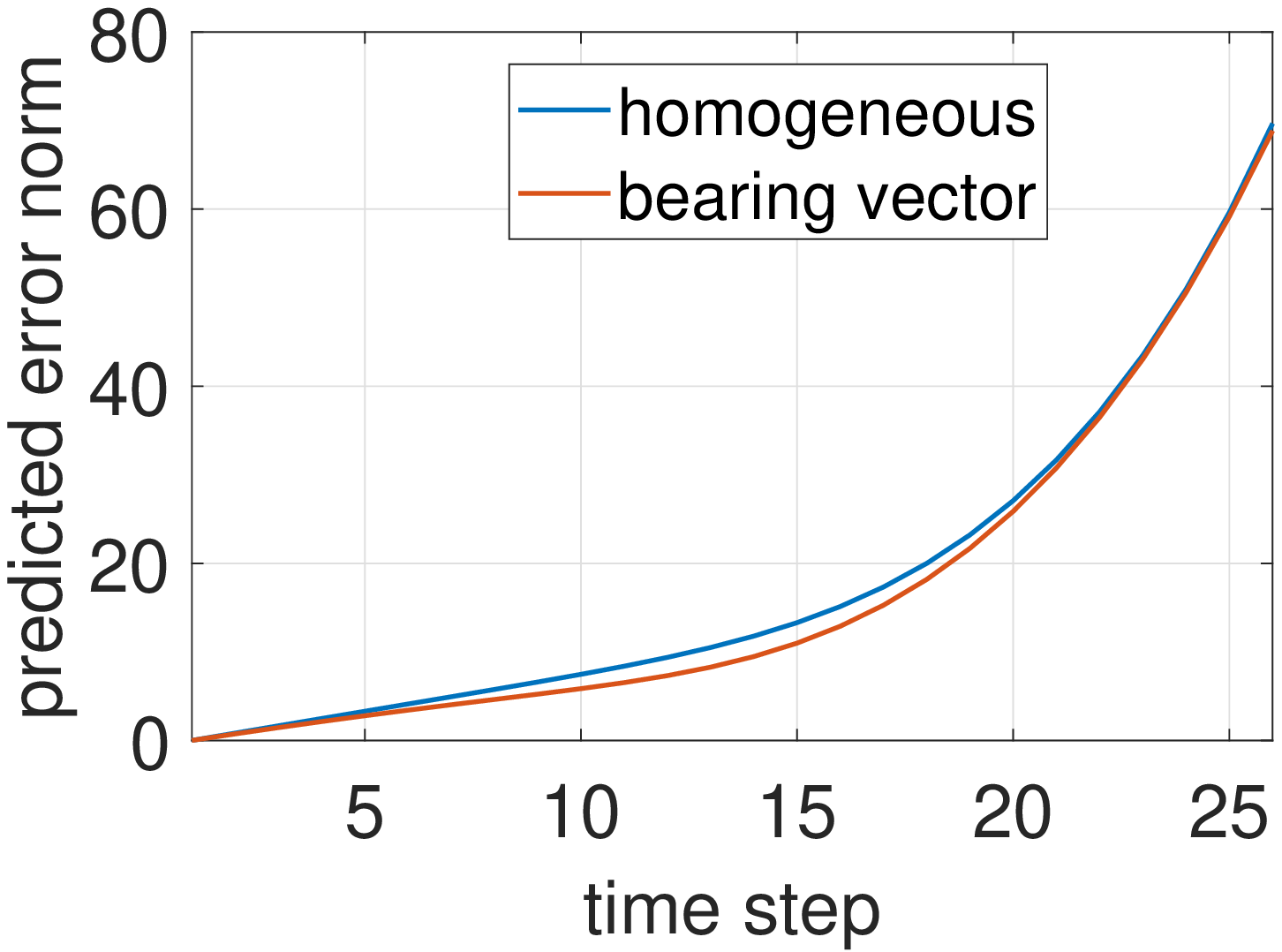}\label{fig_visual_prediction_03}}
	\caption{Comparison of the homogeneous-coordinate-based image dynamics \cite{chaumette2006visual} and the bearing-vector-based image dynamics in visual prediction. (a) Prediction visualization; (b) Norm of the prediction error at each time step.}
	\label{fig_visual_prediction}
\end{figure}

For completeness, we compare the proposed visual prediction strategy with a traditional method \cite{chaumette2006visual}.

We assume the depth and distance are accurate. The predicted features were mapped into a virtual camera for visualization. As shown in Fig. \ref{fig_visual_prediction_02}, the black lines are the actual feature trajectory in the image space, while the green and red lines represent the predicted feature trajectories obtained from a classical homogeneous-coordinate-based method and the bearing-vector-based method, respectively. Fig. \ref{fig_visual_prediction_03} shows that two methods produce similar prediction results which are drifted slowly over time. It indicates that theoretically, the proposed formulation does not improve the prediction accuracy. But considering that depth is more sensitive to orientation changes, we can expect better prediction results from the proposed strategy in real world applications.

\section{Limitations}

The major deficiency of PCVPC is that it cannot work when the drone leaves the $120^{\circ}$ sector region in front of the landmark. As a consequence, it is still challenging to use PCVPC to circle the landmark while keeping the visibility. 

Moreover, PCVPC requires high accuracy of the heading estimate. By controlling the distance, we indeed put the quadrotor at a circle centered at the landmark. If the heading is not accurate, the quadrotor may end up with a wrong place at the circle, where all the perception objectives and constraints can still be satisfied.

\section{Conclusion}\label{Conclusion}

In this work, we present a perception constrained visual predictive control (PCVPC) algorithm for quadrotors to perform agile maneuvers without any localization system. Our framework formulates the image-based visual servoing problem into an optimal control problem in which the quadrotor dynamics, image dynamics, actuation constraints, and visibility constraints are taken into account. To account for depth instability during agile flights, we substitute it by distance and propose a novel formulation for visual servoing. Our method can achieve $9$ m/s in trajectory tracking while keeping the landmark visible. In future work, we will conduct real-world experiments to further validate the proposed algorithm.





\balance
\bibliographystyle{bibtex/bst/IEEEtran}
\bibliography{bibtex/bib/IEEEabrv,bibtex/bib/VPC}

\addtolength{\textheight}{-12cm}   

\end{document}